\def\year{2020}\relax
\documentclass[letterpaper]{article} % DO NOT CHANGE THIS
\usepackage{aaai20}  % DO NOT CHANGE THIS
\usepackage{times}  % DO NOT CHANGE THIS
\usepackage{helvet} % DO NOT CHANGE THIS
\usepackage{courier}  % DO NOT CHANGE THIS
\usepackage[hyphens]{url}  % DO NOT CHANGE THIS
\usepackage{graphicx} % DO NOT CHANGE THIS
\usepackage{graphicx}  % DO NOT CHANGE THIS
\usepackage{multirow}
\usepackage{amsmath}
\usepackage{amsfonts}
\usepackage{bm}
\usepackage{booktabs}
\usepackage{amssymb}
\usepackage{tikz}

\title{Global Context-Aware Progressive Aggregation Network \\
for Salient Object Detection}
\author{Zuyao Chen,\textsuperscript{\rm 1} Qianqian Xu,\textsuperscript{\rm 2} Runmin Cong,\textsuperscript{\rm 3} Qingming Huang\textsuperscript{\rm 1,2,4,5 *}\\
	\textsuperscript{\rm 1}University of Chinese Academy of Sciences, Beijing, China \\
	\textsuperscript{\rm 2}Key Lab. of Intelligent Information Processing, ICT, CAS, Beijing, China \\
	\textsuperscript{\rm 3}Institute of Information Science, Beijing Jiaotong University, Beijing, China \\
	\textsuperscript{\rm 4}Key Lab. of Big Data Mining and Knowledge Management, CAS, Beijing, China\\
	\textsuperscript{\rm 5}Peng Cheng Laboratory, Shenzhen, Guangdong, China \\ 
	chenzuyao17@mails.ucas.ac.cn, xuqianqian@ict.ac.cn, rmcong@bjtu.edu.cn, qmhuang@ucas.ac.cn
}

\def\metrics{$F_\beta\uparrow$\quad$S_m\uparrow$\quad MAE$\downarrow$}
\def\triplets(#1,#2,#3){$#1\quad#2\quad#3$}
 
\begin{document}
\maketitle
\begin{abstract}
Deep convolutional neural networks have achieved competitive performance in salient object detection, in which how to learn effective and comprehensive features plays a critical role. 
Most of the previous works mainly adopted multiple-level feature integration yet ignored the gap between different features. 
Besides, there also exists a dilution process of high-level features as they passed on the top-down pathway.
To remedy these issues, we propose a novel network named \textit{GCPANet} to 
effectively integrate low-level appearance features, high-level semantic features, and global context features through some progressive context-aware 
Feature Interweaved Aggregation (FIA) modules and generate the saliency map in a supervised way. Moreover, a Head Attention (HA) module is used to reduce information redundancy and enhance the top layers features by leveraging the  spatial and channel-wise attention, and the Self Refinement (SR) module is utilized to further refine and heighten the input features.
Furthermore, we design the Global Context Flow (GCF) module to generate the global context information at different stages, which aims to learn the relationship among different salient regions and alleviate the dilution effect of high-level features.
Experimental results on six benchmark datasets demonstrate that the proposed approach outperforms the state-of-the-art methods both quantitatively and qualitatively. 
\end{abstract}
\section{Introduction} 
Salient object detection aims to detect interesting regions that attract human attention in an image \cite{cong2018review}.
As an efficient preprocessing technique, salient object detection benefits a wide range of applications
such as image understanding \cite{zhang2014saliency}, image retrieval \cite{gao2015database}, 
and object tracking \cite{hong2015online}. \\
\begin{figure}[ht]
	\centering 
	\includegraphics[]{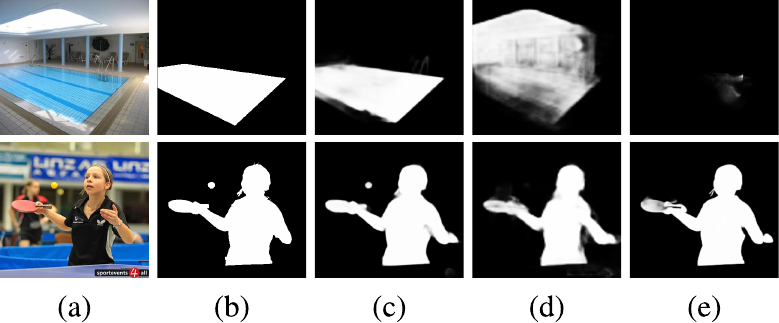}
	\caption {Sample results of our method compared with others. 
		(a) Image; (b) Ground truth; (c) GCPANet (Ours); (d) PiCANet-R \cite{liu2018picanet}; (e) BASNet \cite{qin2019basnet}.
	}
	\label{fig1}
\end{figure} 
\indent In recent years, the development of deep learning, 
especially the emergence of Fully Convolutional Network \cite{long2015fully}, 
has greatly boosted the progress of salient object detection \cite{zhao2015saliency,li2016deep,wang2016saliency}.
Fully Convolutional Network (FCN) stacks multiple convolution layers and pooling layers to gradually enlarge the receptive fields of network
and extracts high-level semantic information.
As pointed out in previous works \cite{luo2017non,zhang2017amulet}, due to the pyramid-like CNNs structure, 
low-level features usually have larger spatial size and more fine-grained details, while high-level features tend to gain more semantic knowledge and discard some meaningless or irrelevant detail information. 
Generally speaking, the high-level features are beneficial to the coarse localization of salient objects,
whereas the low-level features that contain the spatial structural details are suitable to refine boundaries.
However, there remains several problems for the FCN-based methods: 
(1) Due to the gap between different level features, the simple combination of semantic information and appearance information
is insufficient and lacks consideration of the different contribution of different features for salient object detection;
(2) Most of the previous works ignored the global context information, which benefits for deducing the relationship among multiple salient regions and producing more complete saliency result. \\
\indent To remedy the above mentioned issues, we propose a novel network named Global Context-Aware Progressive Aggregation Network (\textit{GCPANet}), which consists of four modules: Feature Interweaved Aggregation (FIA) module, Self Refinement (SR) module, Head Attention (HA) module, and Global Context Flow (GCF) module. Considering the characteristics difference between multiple level features, we design the FIA module to fully integrate the high-level semantic features, low-level detail features, and global context features, which is expected to suppress the noises but recover more structural and detail information. Before the first FIA module, we add a HA module on the top layer of the backbone to strengthen the spatial and channel-wise response on the salient object. After aggregation, features will be fed into a SR module to refine the feature maps via leveraging the inner characteristics within features. Taken into account that the context information can benefit for capturing the relationship among multiple salient objects or different parts of salient object, we design a GCF module to exploit the relationship from global perspective, which is conducive to improving the completeness of salient object detection. Besides, as pointed out in \cite{liu2019simple}, the high-level features will be diluted as they passed on the top-down pathway. By introducing GCF, the features containing global semantics are delivered to feature maps at different stages, which alleviates the effect of features dilution. As shown in Fig. \ref{fig1}, the proposed method can handle some challenging scenarios, such as complex scene understanding (the high-luminance ceiling interference), or multiple objects relationship reasoning (the ping-pong bat and ball). \\
\indent From the above, the contributions of our work can be summarized as follows:
\begin{enumerate} 
\item A global context-aware progressive aggregation network is proposed to achieve saliency detection, which includes the Feature Interweaved Aggregation (FIA) module,
the Self Refinement (SR) module, the Head Attention (HA) module, and the Global Context Flow (GCF) module.
\item The FIA module integrates the low-level detail information, high-level semantic information, and global context information in an interweaved way, where 
the global context information is produced by the GCF module to capture the relationship among different salient regions and improve the completeness of the generated saliency map.
\item Compared with $12$ state-of-the-art methods on six public benchmark datasets, the proposed network \textit{GCPANet} achieves best performance in quantitative and qualitative evaluations.  
\end{enumerate} 
% related work
\section{Related Work}
In this section, we will review the related works on deep learning based salient object detection methods, which have achieved remarkable progress on saliency detection thanks to its powerful representation capability. \\
\indent Inspired by image semantic segmentation, Zhao et al. \cite{zhao2015saliency}
proposed a fully connected CNN to integrate local and global features to predict the 
saliency map. Wang et al. \cite{wang2016saliency} adopted a 
recurrent CNN to refine the predicted saliency map step by step.
For further enhance the saliency map, several recent works 
\cite{hou2017deeply,zhang2017amulet,deng2018r3net,hu2018recurrently,li2018contour,zhang2018bi,zhang2018progressive}
integrate features in multiple layers of CNN to 
exploit the context information at different semantic levels.
Among them, 
Hou et al. \cite{hou2017deeply} introduced short connections to the skip-layer structure 
for capturing fine details.
Zhang et al. \cite{zhang2017amulet} concatenated multi-level feature maps based on multiple resolution 
and introduced a boundary refinement strategy.
Deng et al. \cite{deng2018r3net} proposed an iterative method to optimize the saliency map,
leveraging features generated by deep and shallow layers.
Hu et al. \cite{hu2018recurrently} recurrently concatenated multi-layer features for saliency detection.
Li et al. \cite{li2018contour} proposed a contour-to-saliency transferring method that simultaneously 
predict the contours and saliency maps.
Zhang et al. \cite{zhang2018bi} built a bi-directional message passing model 
for better integrating multi-level features.
Zhang et al. \cite{zhang2018progressive} designed an attention guided network that selectively
integrates multi-level contextual information in a progressive manner. 
Lately, Wu et al. \cite{wu2019cascaded} proposed a cascade partial decoder that utilizes attention mechanism to refine high-level features. 
Qin et al. \cite{qin2019basnet} proposed a boundary-aware model to segment salient object regions and predict the boundaries simultaneously. 
Liu et al. \cite{liu2019simple} extended the FPN  structure equipped with pyramid pooling module 
to fuse the coarse-level semantic features and fine-level features.
% methodology
\section{Methodology}
\begin{figure*}[t]
	\centering
	\includegraphics[]{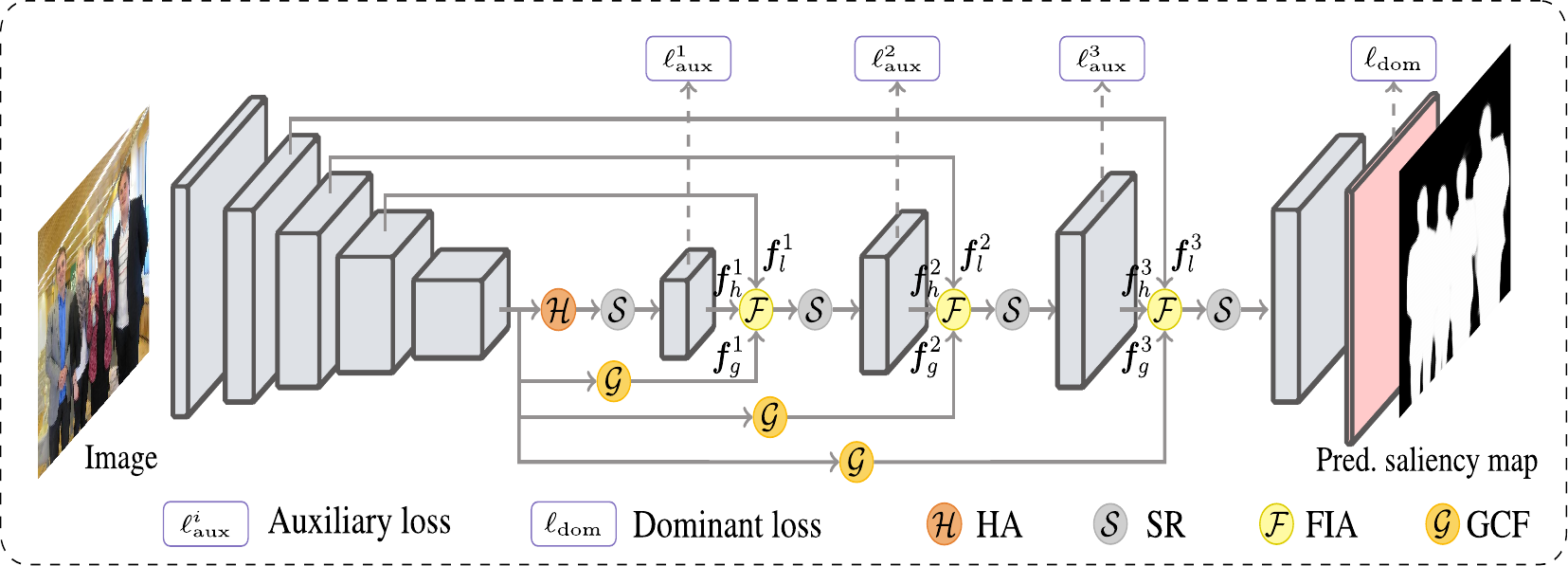}
	\caption{The overall pipeline of the proposed network \textit{GCPANet}. $\bm{f}_l^t$, $\bm{f}_h^t$, $\bm{f}_g^t$ ($t=1,2,3$) denote the low-level detail, high-level semantic, and global context features, respectively.}
	\label{fig:structure}
\end{figure*}
In this section, we first outline the proposed network. Then, we elucidate how each component made up and illustrate its effect for saliency detection.
\subsection{Overview of the Proposed Network}
As Fig. \ref{fig:structure} shows, the proposed network is a symmetrical encoder-decoder architecture, 
where the encoder component is based on ResNet-50 to extract the multi-level features, 
and the decoder component progressively integrates the multi-level comprehensive features to generate the saliency map in a supervised way.
Specifically, we first use a HA module to strengthen the spatial regions and feature channels with high response on salient objects, and a SR module to generate the first-stage high-level features through the feature refinement and enhancement.
Then, we progressively cascade a FIA module and a SR module in three times to learn more discriminative features and generate more accurate saliency map. In the FIA module, the low-level detail information, high-level semantic information, and global context information are fused in an interweaved way. The SR module successive to each FIA module is to refine the coarse aggregation features. Note that, the global context information is produced by the proposed GCF module, 
which captures the relationship among different salient regions and constrains more complete saliency prediction. 
To facilitate the optimization, we combine auxiliary loss branches of each sub-stage with dominant loss.
\subsection{Feature Interweaved Aggregation Module}
As we all know, low-level features include more detail information, such as texture, boundary, and spatial structure, but they also contain more background noises. By contrast, high-level features can provide abstract semantic information, which is beneficial to locate the salient object and suppress the noises. Thus, these two level features are always combined together to generate the complementary features. In addition to these two level features, the global context information is very useful to infer the relationship among different salient objects or parts from the global perspective, which is conducive to generate more complete and accurate saliency map. Moreover, using the context features can alleviate the effect of feature dilution. Hence, we develop the FIA module to fully integrate these three level features, which in turn produces a discriminative and comprehensive feature with global perception. Specifically, as shown in Fig. \ref{fig:cfam}, the FIA module receives three parts input, i.e., the high-level features from the output of the previous layer, the low-level features from the corresponding bottom layer, and the global context feature generated by the GCF module. Note that, the production of global context feature will be introduced in the latter subsection. \\
\indent We first introduce the aggregation strategy for high-level features and low-level features.
Different from previous works \cite{qin2019basnet,liu2019simple} that often simply fuse the high-level features after up-sampling with the low-level features by concatenation or addition operation, 
we adopt a more aggressive yet efficient operation, i.e., multiplication. The multiplication operation can strengthen the response of salient objects, meanwhile suppress the background noises. 
Specifically, for the consistency of multiplication operation, the low-level feature maps $\bm{f}_l^t$ ($t=1,2,3$) are firstly fed into 
a $1\times 1$ convolution layer $conv_1$, which compress the features to have the same number of channels as of the high-level features $\bm{f}_h^t$.
Then, a $3\times 3$ convolution layer is applied to high-level features $\bm{f}_h^t$ to obtain a semantic mask $\bm{W}_{h}^t$ after up-sampling. Further, we multiply the mask $\bm{W}_h^t$ to the compressed low-level features $\tilde{\bm{f}}_l^t$. Besides, considering high-level features will discard some detail information relevant to salient objects, we apply the above fusion strategy in a mirror way. The mirror path different from the above mentioned is that a detail mask $\bm{W}_l^t$ is generated by low-level features through a $3\times 3$ convolution layer and then, the mask $\bm{W}_l^t$ is multiplied to the high-level features $\tilde{\bm{f}_h^t}$ after up-sampling. The mirror path is supposed to add fine-grained detail information to the predicted saliency maps. The above process can be described as 
\begin{figure}[t]
	\centering	
	\resizebox{.45\textwidth}{!}
	{
		\tikzset{every picture/.style={line width=0.75pt}} %set default line width to 0.75pt     
		\begin{tikzpicture}[x=0.75pt,y=0.75pt,yscale=-1,xscale=1]
		\draw (237.5, 0) node   {$\bm{f}_{h}$};
		\draw [->, thick] (237.5,10) -- (237.5,35);
		%rect. 
		\draw   (200,35) -- (275.0,35) -- (275.0,55) -- (200,55) -- cycle ;
		\draw   (237.5, 45) node [align=center] {conv};
		\draw [->, thick] (237.5, 55) -- (237.5, 75);  
		\draw   (200, 75) -- (275.0,75) -- (275.0,95) -- (200, 95) -- cycle ;
		\draw (237.5, 85) node [align=center] {upsample};
		\draw [->, thick] (237.5, 95) -- (237.5, 120);
		\draw (237.5, 128) circle(8);
		\draw [fill=black](237.5, 128)  circle(1);
		%\draw [->, thick] (237.5,136) -- (237.5, 160) ; 
		\draw [->, thick] (237.5,136) -- (237.5, 160) -- (319.5, 187);
		\draw (225, 155) node   {$\bm{f}_{hl}$};
		\draw   (190,105) -- (190,151) -- (170,151) -- (170,105) -- cycle ;
		\draw   (180, 128) node [rotate=-90] [align=left] {conv};
		\draw  [->, thick]   (190,128) -- (229.5,128) ;
		\draw  [->, thick]   (136.42,128) -- (170,128) ;
		\draw (206,118) node   {$\tilde{\bm{f}}_{l}$}; 
		\draw (128,128) node   {$\bm{f}_{l}$};
		\draw   (290, 75) -- (365, 75) -- (365, 95) -- (290, 95) -- cycle ;
		\draw   (327.5, 85) node  [align=center] {upsample};
		%Curve Lines  
		\draw   [->, thick] (238,23) .. controls (334.71,23) and (327.5, 60) .. (327.5, 75) ;
		\draw   [->, thick] (327.5, 95) -- (327.5, 120) ;
		\draw (327.5, 128) circle(8);
		\draw [fill=black](327.5, 128)  circle(1);
		\draw [->, thick] (327.5,136) -- (327.5, 182) ;
		\draw (315, 155) node   {$\bm{f}_{lh}$};
		\draw   (380,105) -- (380, 151) -- (360,151) -- (360,105) -- cycle ;
		\draw  [->, thick] (360, 128) -- (335.5,128) ;
		\draw  [->, thick] (400, 128) -- (380, 128) ;
		\draw (407, 128) node   {$\tilde{\bm{f}}_{l}$};
		\draw (370, 128) node [rotate=-90] [align=center] {conv};
		\draw (450, 0) node   {$\bm{f}_{g}$};
		\draw [->, thick] (450, 10) -- (450, 35) ;
		\draw (410, 35) -- (485, 35) -- (485, 55) -- (410, 55) -- cycle ;
		\draw (450, 45) node  [align=center] {conv};
		\draw [->, thick] (450, 55) -- (450, 75) ;
		\draw   (410, 75) -- (485, 75) -- (485, 95) -- (410,95) -- cycle ;
		\draw (450, 85) node  [align=center] {upsample};
		\draw (450, 128) circle(8);
		\draw [fill=black](450, 128) circle(1);
		\draw [->, thick] (450,95) -- (450, 120) ;
		\draw [->, thick] (450, 136) -- (450, 160) -- (335.5, 187);
		\draw [->, thick] (500,128) -- (458,128) ;
		\draw (462.5,155) node   {$\bm{f}_{gl}$};
		\draw (505, 128) node   {$\tilde{\bm{f}}_{l}$}; 
		
		\draw (327.5, 190) circle(8);
		\draw  (327.5, 190) node {c};
		\draw [->, thick] (327.5, 198) -- (327.5, 210);
		\draw (290, 210) -- (365, 210) -- (365, 230) -- (290, 230) -- cycle;
		\draw (327.5, 220) node [align=center] {conv};
		\draw [->, thick](327.5, 230) -- (327.5, 245);
		\draw (327.5, 255) node {$\bm{f}_{a}$};
		% end of tikzpicture
		\end{tikzpicture}
		
	}
	\caption{Illustration of the FIA module, where symbol ``c'' denotes concatenation.}
	\label{fig:cfam}
\end{figure}
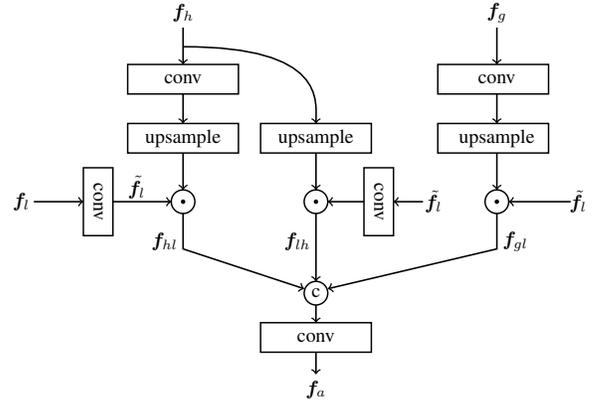
\begin{align}
&\bm{W}_h^t = upsample(conv_2(\bm{f}_h^t)) \\
&\bm{f}_{hl}^t = \delta(\bm{W}_h^t \odot \tilde{\bm{f}_l^t}) \\
&\bm{W}_l^t = conv_3(\tilde{\bm{f}_l^t}) \\
&\bm{f}_{lh}^t = \delta(\bm{W}_l^t \odot upsample(\bm{f}_h^t))  
\end{align}
where $\tilde{\bm{f}}_l^t = conv_1(\bm{f}_l^t)$ denotes the compressed low-level features, $\odot$ denotes element-wise multiplication, $\delta$ denotes the ReLU activation function, $upsample$ is the up-sampling operation via bilinear interpolation, and $t$ is the stage index. \\
\indent Further, to model the relationship between different parts of salient objects and alleviate the dilution process of high-level features, we introduce the global context features $\bm{f}_g^t$ at each stage. We employ the global context features $\bm{f}_g^t$ to generate a context mask 
$\bm{W}_g^t$. Then, the mask $\bm{W}_g^t$ is multiplied to the compressed low-level features $\tilde{\bm{f}_l^t}$. 
\begin{align}
 &\bm{W}_g^t = upsample(conv_4(\bm{f}_g^t)) \\
 &\bm{f}_{gl}^t = \delta(\bm{W}_g^t \odot \tilde{\bm{f}_l^t})
\end{align}
Finally, these three level features are concatenated and then passed through a $3\times 3$ convolution layer to 
obtain the final fusion features:
\begin{align}
	\bm{f}_{a}^t = conv_5(concat(\bm{f}_{hl}^t, \bm{f}_{lh}^t, \bm{f}_{gl}^t))
\end{align}
Each of the above mentioned convolution layers except $conv_2$, $conv_3$, and $conv_4$ is equipped with a batch normalization layer and the ReLU activation function.
The output of FIA module is then passed to the SR module.
\subsection{Self Refinement Module}
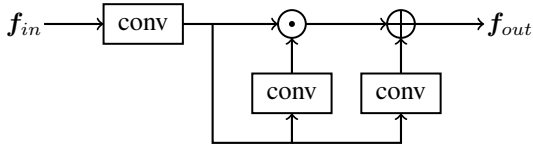
\begin{figure}[t] 
\centering 
\tikzset{every picture/.style={line width=0.75pt}} %set default line width to 0.75pt        
\begin{tikzpicture}[x=0.75pt,y=0.75pt,yscale=-1,xscale=1]
%box 1
\draw   (40,60) -- (80,60) -- (80,80) -- (40,80) -- cycle ;
\draw (60, 70) node  [align=center] {conv};
%box 2 
\draw   (115,95) -- (155,95) -- (155,115) -- (115,115) -- cycle ;
\draw (135,105) node  [align=center] {conv}; 
%box 3
\draw   (170,95) -- (210,95) -- (210,115) -- (170,115) -- cycle ;
\draw (190,105) node  [align=center] {conv};
%odot
\draw (135, 70) circle(7);
\draw [fill=black](135, 70) circle(1);
%hline 1
\draw    [->, thick](10, 70) -- (40,70) ;
%oplus 
\draw (190, 70) circle(7);
\draw (183, 70) -- (197, 70);
\draw (190, 63) -- (190, 77);
%hline 2
\draw [->, thick](80, 70) -- (128, 70); 
%hline 3
\draw [->, thick] (142, 70) -- (183, 70);
%hline 4
\draw [->, thick] (197,70) -- (232, 70) ;
%vline1
\draw  [->, thick] (135, 95) -- (135,77) ;
%vline 2
\draw  [->, thick] (190,95) -- (190, 77) ;
%text nodes
\draw (245,70) node   {$\bm{f}_{out}$};
\draw (0, 70) node   {$\bm{f}_{in}$};
%vline 3,4
\draw [->, thick](95, 70) -- (95, 130) -- (135, 130)
	 -- (135, 115);
\draw [->, thick] (135, 130) -- (190, 130) -- (190, 115);
\end{tikzpicture}
\caption{Structure of the SR module.}
\label{fig:srm}
\end{figure} 
In FIA module, we combine the complementary characteristics between different level features and obtain the comprehensive feature expression. 
As a simple and intuitionistic way, one can directly apply a softmax layer after FIA module to obtain the saliency maps, while it still exists some defects. For instance, 
there are some holes in the predicted salient objects, which are caused by the contradictory response of different layers. 
Hence, we develop a SR module to further refine and enhance the feature maps after passing the HA module and FIA modules by utilizing the multiplication and addition operation
(see Fig. \ref{fig:srm}).
In detail, we firstly apply a $3\times 3$ convolution layer to squeeze the input features $\bm{f}_{in}$
into feature vector $\tilde{\bm{f}}$ with the channel dimension of $256$, meanwhile remaining useful information. 
Then, the feature $\tilde{\bm{f}}$ is fed into two convolution layers to obtain the mask $\bm{W}$ and bias $\bm{b}$ for multiplication and addition operation.
The main process can be described as
\begin{align}
	&\tilde{\bm{f}} = conv_6 (\bm{f}_{in}) \\
	%&(\bm{W}, \bm{b}) = conv_7 (\tilde{\bm{f}}) \\
	&\bm{f}_{out} = \delta(\bm{W}\odot \tilde{\bm{f}} + \bm{b})
\end{align}
where $\bm{f}_{out}$ is the refined feature maps.
\subsection{Head Attention Module} 
Since the top layers features of the encoder component usually are redundant for salient object detection, we design a HA module following the top layer to learn more selective and representative features by leveraging the spatial and channel-wise attention mechanisms. \\
\indent Specifically, we first apply a convolution layer to the input feature maps ${\bm{F}}$ to 
obtain a compressed feature representation $\tilde{\bm{F}}$ with $256$ channels. 
Then, we generate a mask $\bm{W}$ and bias $\bm{b}$ as similar as the way used in the SR module. 
The output of the first stage is obtained by
\begin{align}
%&\tilde{\bm{F}} = conv_8(\bm{F}) \\
%&(\bm{W}, \bm{b}) = conv_9(\bm{F}) \\
&\bm{F}_1 = \delta(\bm{W}\odot \tilde{\bm{F}} + \bm{b}) 
\end{align}
\indent Further, the input feature $\bm{F}$ is down-sampled into
a channel-wise feature vector ${\bm{f}}$ through average pooling, which has strong consistency and invariance.
Then, two successive fully connected layers $fc_1(\cdot), fc_2(\cdot)$ are applied to 
project the feature vector $\bm{f}$ into an output vector $\bm{y}$.
The final output feature maps $\bm{F}_{out}$ will be obtained via weighting with vector $\bm{y}$.
The second stage can be described as the following equations,
\begin{align}
	%&\bm{f} = \frac{1}{H\times W}\sum_{y=1}^{H}\sum_{x=1}^{W} \bm{F}(x, y) \\
	&\bm{y} = \sigma\circ fc_2\circ \delta\circ fc_1\circ (\bm{f}) \label{eq:att} \\
	&\bm{F}_{out} = \bm{F}_1\odot \bm{y} 
\end{align}
where $fc_i(\cdot)$ denotes $i$-th FC layers, $\delta$ denotes the ReLU activation function, $\sigma$ is the sigmoid operation,
and $\circ$ denotes function composition.
\subsection{Global Context Flow Module}
For the challenging scenarios in salient object detection, such as cluttered background, foreground disturbance, and multiple salient objects, simple integration of high-level and low-level features may fail to completely detect the salient regions due to lacking the global semantic relationship among different parts of salient object or multiple salient objects. 
Besides, since the top-down pathway is built upon the bottom-up backbone, the high-level features will be gradually diluted as they are transmitted to lower layers.  \\
\indent To remedy these issues, we design the GCF module to capture the global context information embedded into the FIA module at each stage.
Different from \cite{liu2019simple}, we take into account the different contributions at different stages.
We firstly employ global average pooling \cite{lin2013network} to obtain the global contextual information and then reassign different weights to different channels of the global contextual feature maps for each stage.
More specifically, for each stage, the process can be described as 
\begin{align}
	%&\bm{y}^t = \sigma(fc_2(\delta(fc_1(\bm{f}_g^t)))) \\
	&\bm{y}^t = \sigma\circ fc_4 \circ \delta\circ fc_3(\bm{f}_{gap}) \\ 
	& \tilde{\bm{f}}^t = conv_{10}(\bm{f}_{top}) \\
	&\bm{f}^t_g = \tilde{\bm{f}}^t\odot \bm{y}^t
\end{align}
where $\bm{f}_{top}$ refers to the top layer features, and $\bm{f}_{gap}$ refers to the features generated by the top layer features through global average pooling, which includes global contextual information. Then, the output ${\bm{f}_g^t}$ is fed into the FIA module, which has been elaborated in the previous section.
\subsection{Loss Function}
In saliency detection, binary cross-entropy loss is often used as the loss function to measure the relation between the generated saliency map and 
the ground truth, whcih can be formulated as
\begin{equation}
\begin{split}
\ell & = -\frac{1}{H\times W} \sum_{i=1}^{H}\sum_{j=1}^{W}[G_{ij}\log(S_{ij})  \\
& + (1-G_{ij})\log(1 - S_{ij})] 
\end{split} 
\end{equation}
where $H$, $W$ denote the height and width of the image, respectively, 
$G_{ij}$ is the ground truth label of the pixel $(i,j)$, and $S_{ij}$ 
represents the corresponding probability of being salient objects in position $(i,j)$. To facilitate the optimization of the proposed network, 
we add auxiliary loss at three decoder stages. Specifically, a $3\times 3$ convolution operation is applied 
for each stage to squeeze the channel of the output feature maps to $1$.
Then these maps are up-sampled to the same size as the ground truth via bilinear interpolation and sigmoid function is used to normalize the predicted values into $[0,1]$.
The total loss consists of two parts, i.e., the dominant loss corresponding to the output
and the auxiliary loss of each sub-stage.
%The whole loss combines all the losses together, which is defined as 
\begin{equation}
 \ell_{\mathrm{total}} = \ell_{\mathrm{dom}} + \sum_{i=1}^{3} \lambda_{i}\ell_{\mathrm{aux}}^i
\end{equation}
where $\lambda_i$ denotes the weight of different loss, and $\ell_{\mathrm{dom}}$, 
$\ell_{\mathrm{aux}}^i$ denote the dominant and auxiliary loss, respectively.
The auxiliary loss branches only exist during the training stage, whereas they are abandoned when inference.
% experiments
\begin{table*}[t]
	\centering
	\caption{Performance comparison with $12$ state-of-the-art methods on $6$ benchmark datasets. 
		The best results on each dataset are highlighted in \textbf{boldface}.}
	\resizebox{\textwidth}{0.12\textheight}
	{
		\begin{tabular}{|c|c|c|c|c|c|c|}
			\hline
			\multirow{2}{*}{Methods} & ECSSD & HKU-IS & PASCAL-S & DUT-OMRON & DUTS-TE & SOD\\  \cline{2-7} 
			& \metrics & \metrics & \metrics  & \metrics  & \metrics  & \metrics  \\ \hline 
			Amulet \cite{zhang2017amulet} & \triplets (0.915,0.894, 0.059) & \triplets (0.894, 0.882, 0.053) & \triplets (0.832,0.815,0.097)
			&\triplets (0.744,0.781,0.097) &\triplets (0.779,0.803,0.085) &\triplets (0.803,0.754,0.139) \\ \hline 
			C2SNet \cite{li2018contour} & \triplets(0.911,0.896,0.053) & \triplets(0.898, 0.888, 0.046) & \triplets(0.848,0.834,0.080)
			&\triplets(0.759,0.799,0.072) &\triplets(0.811,0.831,0.062) &\triplets (0.819,0.757,0.121) \\ \hline 
			RADF \cite{hu2018recurrently} & \triplets (0.912,0.888,0.064) & \triplets (0.903,0.882,0.052) & \triplets (0.832,0.804,0.104)
			&\triplets (0.785,0.811,0.071) &\triplets (0.811,0.820,0.073) & \triplets (0.822,0.756,0.136) \\ \hline 			
			RANet \cite{chen2018reverse} & \triplets (0.920,0.894,0.055) & \triplets (0.912,0.888,0.045) &\triplets (0.830,0.792,0.102)
			&\triplets (0.785,0.812,0.063) &\triplets (0.831,0.839,0.060) & \triplets (0.847,0.761,0.122) \\ \hline 
			DGRL \cite{wang2018detect} & \triplets(0.925,0.906,0.043) &\triplets(0.913,0.896,0.037) &\triplets(0.853,0.834,0.074)
				&\triplets(0.779,0.810,0.063) &\triplets(0.834,0.846,0.051) &\triplets(0.844, 0.770, 0.104) \\ \hline 
			PAGR \cite{zhang2018progressive} & \triplets (0.926,0.889,0.061) &\triplets (0.918,0.887,0.048) &\triplets (0.851,0.813,0.092)
			&\triplets (0.771,0.775,0.071) &\triplets (0.854,0.838,0.056) &\triplets (0.836,0.714,0.145) \\ \hline
			$\text{R}^3$Net \cite{deng2018r3net} & \triplets(0.929,0.910,0.051) & \triplets (0.910,0.894,0.047) &\triplets(0.837,0.809,0.101)
			&\triplets (0.793,0.819,0.073) &\triplets (0.829,0.837,0.067) &\triplets (0.837,0.765,0.129) \\ \hline 
			BMPM \cite{zhang2018bi} & \triplets(0.929,0.911,0.044) & \triplets (0.920,0.906,0.039) &\triplets (0.857,0.840,0.073)
			&\triplets (0.775,0.809,0.063) &\triplets (0.852,0.862,0.049) & \triplets (0.852,0.784,0.105) \\ \hline 
			%PFANet \cite{zhao2019pyramid} & \triplets (0.922,0.871,0.045) & \triplets (0.926,0.876,0.032) &\triplets(0.875, 0.798, {0.063})
			%&\triplets ({0.856},{0.774}, {0.042}) &\triplets ({0.870},{0.786},{0.041}) & - \\ \hline 	
			PiCANet-R \cite{liu2018picanet} & \triplets(0.935,0.917,0.046) & \triplets(0.919,0.904,0.043) &\triplets(0.863,0.849, 0.075)
				&\triplets (0.803,0.832,0.065) &\triplets (0.860,0.869,0.051) &\triplets (0.853,0.787,0.102) \\ \hline	
			CPD-R \cite{wu2019cascaded} &  \triplets (0.939,0.918,0.037) & \triplets (0.925,0.906,0.034) &\triplets (0.864,0.842,0.072)
			&\triplets (0.797,0.825,0.056) &\triplets (0.865, 0.869, 0.043) & \triplets (0.857,0.765,0.110) \\ \hline 
			BASNet \cite{qin2019basnet} &\triplets(0.943,0.916,0.037) & \triplets (0.928,0.909, 0.032) &\triplets (0.857,0.832,0.076)
			&\triplets (0.805,0.836,0.057) & \triplets(0.859,0.866,  0.048) & \triplets (0.849, 0.766, 0.112) \\ \hline 
			PoolNet \cite{liu2019simple} & \triplets (0.944, 0.921,0.039) & \triplets (0.933, 0.917, 0.032) & \triplets (0.869, 0.845, 0.074)
			&\triplets (0.808, 0.836, 0.056) &\triplets (0.880,0.883,0.040) &\triplets (0.867, 0.795, 0.100) \\ \hline 
			\textbf{Ours} & \triplets(\textbf{0.949}, \textbf{0.927}, \textbf{0.035}) &\triplets (\textbf{0.938},\textbf{0.920},\textbf{0.031})&\triplets (\textbf{0.876},\textbf{0.861},\textbf{0.061}) 
			&\triplets (\textbf{0.812},\textbf{0.839},\textbf{0.056}) & \triplets(\textbf{0.888}, \textbf{0.891}, \textbf{0.038}) & \triplets (\textbf{0.872},\textbf{0.802}, \textbf{0.087}) \\ \hline 
			%\textbf{Ours (ResNet34)} & \triplets(0.946, 0.927, 0.036) &\triplets (0.937, 0.921, 0.031) & \triplets (0.876,0.856,0.065) &\triplets (0.812, 0.842, 0.057) & \triplets (0.888,0.890,0.040) &\triplets (0.872, 0.805,0.089) \\ \hline
			%\textbf{$\text{Ours}^{*}$} & \triplets(0.947, 0.927, 0.036) &\triplets (0.936, 0.919, 0.032) & \triplets (0.873,0.858,0.064) &\triplets (0.812, 0.837, 0.057) & \triplets (0.888,0.889,0.039) &\triplets (0.872, 0.802,0.090) \\ \hline			
		\end{tabular}} 
		\label{table:pf}
	\end{table*} 
	\begin{figure*}[ht]
		\centering 
		\includegraphics[width=\textwidth, height=0.27\textheight]{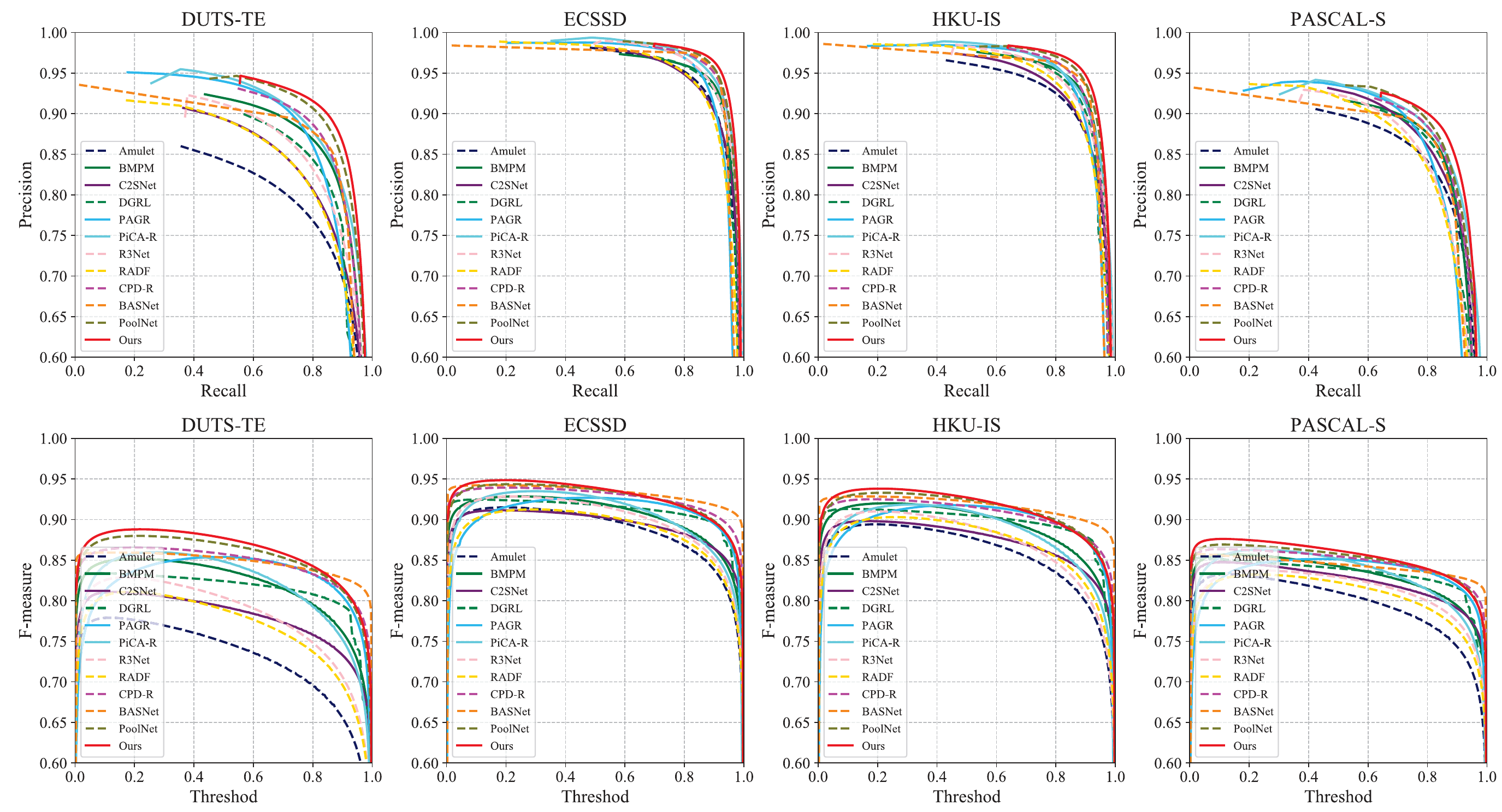}
		\caption{Illustration of PR curves (the first row), F-measure curves (the second row) on four datasets.}
		\label{fig:pr}
	\end{figure*}
\section{Experiments}
\begin{figure*}[ht]
	\centering
	\includegraphics[]{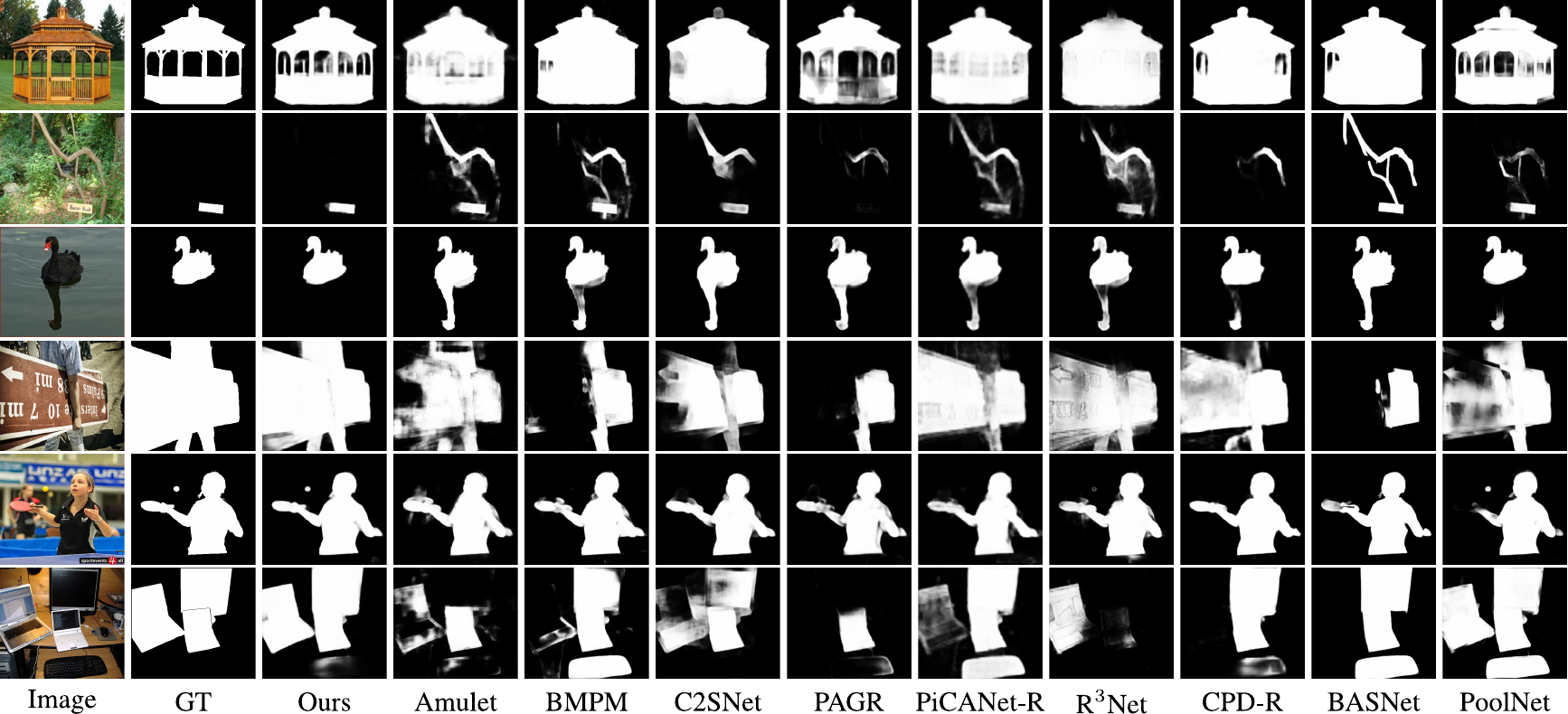}
	\caption{Qualitative comparison of the proposed model with other state-of-the-art methods. Obviously, saliency maps generated by our approach are
		more accurate and much close to the ground truth in various challenging scenarios.}
	\label{fig:vis}
\end{figure*}
In this section, we first describe the implementation details, introduce the benchmark datasets, evaluation metrics.
Then, we conduct experiments on these datasets to evaluate the effectiveness of the proposed method.
\subsubsection{Implementation Details}
We adopt ResNet-50 \cite{he2016deep} pretrained on ImageNet \cite{deng2009imagenet} as our network backbone. 
In the training stage, we resize each image to $320\times 320$ with random horizontal flipping,
then randomly crop a patch with the size of $288\times 288$ for training. During the inference stage, 
images are simply resized to $320\times 320$ then fed into the network to obtain prediction without any other post-processing (e.g., CRF).
We use Pytorch \cite{paszke2017automatic} to implement our model. 
Mini-batch Stochastic gradient descent (SGD) is used to optimize the whole network 
with the batch size of $32$, the momentum of $0.9$, and 
the weight decay of $5e\textit{-}4$. 
We use the warm-up and linear decay strategies with the maximum learning rate $5e\textit{-}3$ for the
backbone and $0.05$ for other parts to 
train our model and stop training after $30$ epochs.
The inference of a $320\times 320$ image takes about $0.02$s (over $50$ fps) with the acceleration of one NVIDIA Titan-Xp GPU card. The code is now available. \footnote{https://github.com/JosephChenHub/GCPANet.git}
\subsubsection{Datasets} 
We conduct experiments on six public saliency detection
benchmark datasets, and the detailed introduction is provided as follows:
\begin{itemize}
\item \textit{ECSSD} \cite{yan2013hierarchical} consists of 1,000 natural images which are manually collected from the Internet; 
\item \textit{PASCAL-S} \cite{li2014secrets} has 850 natural images that are carefully selected from the PASCAL VOC dataset \cite{everingham2010pascal}; 
\item \textit{HKU-IS} \cite{li2015visual} includes 4,447 images and most of them have low contrast or more than one salient object; 
\item \textit{DUT-OMRON} \cite{yang2013saliency} contains 5,168 high quality images. Images of this dataset have one or more salient objects and relatively cluttered background, and thus salient object detection on this dataset is very challenging; 
\item \textit{SOD} \cite{movahedi2010design} is composed of 300 images, many of which contain multiple objects either with low contrast or touching the image boundary; 
\item \textit{DUTS} \cite{wang2017learning} is currently the largest saliency detection benchmark dataset,
which consists of 10,553 training images (\textit{DUTS-TR}) and 5,019 testing images (\textit{DUTS-TE}).
\end{itemize}
As with other works in salient object detection \cite{qin2019basnet,liu2019simple}, we employ \textit{DUTS-TR} as our training dataset and evaluate our model on other datasets. 
\subsubsection{Evaluation Metrics} 
To quantitatively evaluate the effectiveness of our proposed model, we adopt precision-recall (PR) curves , F-measure ($F_\beta$) score and curves, Mean Absolute Error (MAE), and
structural similarity measure ($S_m$) as our performance measures. 
With different thresholds, pairs of precision and recall value can be computed by comparing the binarized map with the ground truth. 
Then, we can plot the precision-recall curve \cite{cong2019video}. 
The second metric F-measure score takes both precision and recall into account,
which is defined as 
$
  F_{\beta} = \frac{(1+{\beta}^2)\cdot{Precision}\cdot{Recall}}{{\beta}^2\cdot Precision + Recall}
$
where ${\beta}^2$ is set to $0.3$ to emphasize the precision over recall,
as suggested in the previous work \cite{cong2018hscs}.
Larger F-measure score indicates better performance.
For precision-recall pairs, we calculated each corresponding F-measure score and choose the maximum as the evaluation score on the whole dataset.
Another metric MAE is defined as the average pixel-wise absolute difference
between the prediction map and the ground truth \cite{cong2017iterative}, i.e., 
$
  MAE = \frac{1}{H\times W}\sum_{y=1}^{H}\sum_{x=1}^{W}|S(x, y) - G(x, y)|
$
where $S$ denotes the predicted saliency map, $G$ indicates the corresponding 
ground truth, and $H$, $W$ are the height and width of the saliency map respectively.
The smaller MAE indicates better performance.
Since the $F_{\beta}$ and MAE are based on pixel-wise errors and ignore the structural similarities, 
we adopt the structural similarity measure proposed by \cite{fan2017structure} as one of our metrics.
The structural similarity measure is defined as 
$ S_m = \alpha * S_o + (1-\alpha) * S_r $, where $\alpha$ is set to $0.5$ to balance the object-aware structural similarity ($S_o$) and region-aware structural similarity ($S_r$). 
\subsection{Compared with the State-of-the-arts}
We compare the proposed model with $12$ state-of-the-art methods, including Amulet \cite{zhang2017amulet}, 
C2S \cite{li2018contour}, 
RANet \cite{chen2018reverse}, PAGR \cite{zhang2018progressive},
PiCANet-R \cite{liu2018picanet}, DGRL \cite{wang2018detect}, $\text{R}^3$Net \cite{deng2018r3net}, 
BMPM \cite{zhang2018bi}, RADF \cite{hu2018recurrently}, CPD-R \cite{wu2019cascaded}, 
BASNet \cite{qin2019basnet} and PoolNet \cite{liu2019simple}.
For fair comparison, the saliency maps of different methods are provided by authors 
or obtained by running their released codes under the default parameters.
\subsubsection{Quantitative Evaluation}
Table \ref{table:pf} shows the quantitative comparison results in terms of F-measure, S-measure, and MAE score.  
It's obvious that the proposed method achieves the best performance in terms of different measures, 
which demonstrates the effectiveness of the proposed model. 
In addition, as shown in Fig. \ref{fig:pr}, the PR curves and F-measure curves by our approach (the red curves) are outstanding in most cases compared with
other previous methods under different thresholds, which is consistent with the measures reported in Table \ref{table:pf}.
\subsubsection{Qualitative Evaluation}
To further illustrate the advantages of the proposed method, 
we provide some visual examples of different methods.
%As Fig. \ref{fig:vis} shows, our proposed model can handle various challenging scenarios, including certain manufactured structures (the first row), cluttered background (the second row), low contrast foregrounds (the third row), large objects (the fourth row), objects concurrency (the fifth row) and multiple salient objects (the sixth row).
As Fig. \ref{fig:vis} shows, our proposed method can handle various challenging scenarios, including fine-grained structures,
cluttered background, foreground disturbance, objects concurrency, and multiple salient objects, etc.
Compared with other previous methods, the saliency maps generated by our approach are more complete and accurate. Note that our approach is more robust to background/foreground disturbance (the second/third row) and can capture the relationship among multiple objects (the fifth row), which illustrates the power of the feature interweaved aggregation strategy and the introducing of global context information.
\subsection{Ablation Study}
\begin{table*}[ht]
	\caption{MAE Comparison of the GCF with the shared one.}
	\centering
	%\resizebox{.45\textwidth}{!} 
	{
		\begin{tabular}{ccccccc}
			\toprule[2pt]
			& ECSSD &HKU-IS & PASCAL-S & DUT-OMRON &DUTS-TE & SOD\\
			\midrule[1pt]
			with the Shared & 0.0361& 0.0313 &0.0628& 0.0590 & 0.0388 & 0.0915 \\	
			with \textbf{GCF}  & \textbf{0.0348} &\textbf{0.0309}& \textbf{0.0614} &  \textbf{0.0563} & \textbf{0.0380} & \textbf{0.0874}\\
			\bottomrule[2pt]
		\end{tabular}
	}
	\label{tab:gcfm}
\end{table*}
\begin{table}[ht]
	\caption{Ablation study with different components combinations on ECSSD dataset.}
	\label{tab:ablation}
	\centering
	\resizebox{0.46\textwidth}{!}{
		\begin{tabular}{ccccc|c}
			\toprule[2pt]
			Baseline & FIA & SR & HA & GCF &$MAE\downarrow$\\
			\midrule[1pt]
			\checkmark & & & &  &0.0456\\
			\checkmark & \checkmark & & &  &0.0390\\
			\checkmark & \checkmark & \checkmark &    &&0.0365\\
			\checkmark & \checkmark & \checkmark & \checkmark & &0.0364 \\
			%\checkmark & \checkmark & \checkmark &  &\checkmark& 0.0350 \\
			\checkmark & \checkmark & \checkmark & \checkmark & \checkmark & \textbf{0.0348} \\
			\bottomrule[2pt]
		\end{tabular}
	}
\end{table}
In this part, we conduct the ablation study to verify the effectiveness 
of each key components designed in the proposed model. 
The ablation experiments are conducted on the ECSSD dataset and ResNet-50 is adopted as the backbone. 
As shown in Table \ref{tab:ablation}, the proposed model containing all components (i.e., FIA, SR, HA and GCF) achieves the best performance,
which demonstrates the necessity of each component for the proposed model to obtain the best saliency detection results.\\
\indent We adopt the model like U-Net \cite{ronneberger2015u} that only concatenates high-level features after up-sampling and low-level features as the baseline model, then add each module progressively. From Table \ref{tab:ablation},  
the FIA module largely improves the baseline from $0.0456$ to $0.0390$ in terms of MAE.
Furthermore, the MAE score is improved by $14\%$ compared with the basic model after adding the SR module.
The combination of FIA and SR has already achieved well performance, while the addition of HA has a slight enhancement. 
Finally, we add the GCF to the model and obtain the best result. \\
\indent Moreover, we evaluate the effectiveness of the GCF module compared to another setting, 
in which the global context features are shared at all stages.
From Table \ref{tab:gcfm}, the proposed GCF module outperforms the shared one.
The potential reason behind this phenomenon is that the parallel scheme of the GCF modules can provide distinct features for different stages,
which benefits to learn the comprehensive and discriminative features for salient objects. 
% conclusion
\section{Conclusion}
In this paper, we propose a Global Context-Aware Progressive Aggregation Network (GCPANet) to achieve salient object detection.
Considering different characteristics of different level features,  we design a simple yet effective aggregation module to fully integrate different level features. We introduce global context information at different stages to capture the relationship among multiple salient objects or multiple regions of salient object and alleviate the dilution effect of features.
Experimental results on six benchmark datasets demonstrate that the proposed network outperforms other 12 state-of-the-art methods
under different evaluation metrics.
\section{Acknowledgement}
This work was supported in part by National Natural Science Foundation of China: 61620106009, 61931008, U1636214, 61836002, 61672514 and 61976202, in part by Key Research Program of Frontier Sciences, CAS: QYZDJ-SSW-SYS013, in part by Beijing Natural Science Foundation (No. 4182079), in part by the Strategic Priority Research Program of Chinese Academy of Sciences, Grant No. XDB28000000, in part by the Fundamental Research Funds for the Central Universities under Grant 2019RC039, and in part by Youth Innovation Promotion Association CAS.

% References and End of Paper
% These lines must be placed at the end of your paper
\bibliography{Bibliography-File}
\bibliographystyle{aaai}
\end{document}